\documentclass[11pt]{article}
\usepackage{amssymb}
\usepackage[preprint]{acl}

\usepackage{times}
\usepackage{latexsym}
\usepackage[T1]{fontenc}
\usepackage[utf8]{inputenc}
\usepackage{microtype}
\usepackage{inconsolata}
\usepackage{graphicx}
\usepackage{amsmath}
\usepackage{booktabs}
\usepackage{booktabs,subcaption,makecell}

\newcommand{\method}{\textsc{LaMI }} 
\title{LaMI: Augmenting Large Language Models via Late Multi-Image Fusion}

\author{
  Guy Yariv\textsuperscript{1} \quad
  Idan Schwartz\textsuperscript{2}\thanks{\hspace{1.5mm}Equal contribution.} \quad
  Yossi Adi\textsuperscript{1}\footnotemark[1] \quad
  Sagie Benaim\textsuperscript{1}\footnotemark[1] \\ 
  \textsuperscript{1}The Hebrew University of Jerusalem \quad
  \textsuperscript{2}Bar-Ilan University \\
  \texttt{guy.yariv@mail.huji.ac.il}
}

\begin{document}

\maketitle

\begin{abstract}
Commonsense reasoning often requires both textual and visual knowledge, yet Large Language Models (LLMs) trained solely on text lack visual grounding (e.g., ``what color is an emperor penguin's belly?''). Visual Language Models (VLMs) perform better on visually grounded tasks but face two limitations: (i) often reduced performance on text-only commonsense reasoning compared to text-trained LLMs, and (ii) adapting newly released LLMs to vision input typically requires costly multimodal training. An alternative augments LLMs with test-time visual signals, improving visual commonsense without harming textual reasoning, but prior designs often rely on early fusion and a single image, which can be suboptimal. We propose a \emph{late multi-image fusion} method: multiple images are generated from the text prompt with a lightweight parallel sampling, and their prediction probabilities are combined with those of a text-only LLM through a late-fusion layer that integrates projected visual features just before the final prediction. Across visual commonsense and NLP benchmarks, our method significantly outperforms augmented LLMs on visual reasoning, matches VLMs on vision-based tasks, and, when applied to strong LLMs such as LLaMA~3, also improves NLP performance while adding only modest test-time overhead. Project page is available at: \url{https://guyyariv.github.io/LaMI/}.
\end{abstract}

\section{Introduction}

Large Language Models (LLMs) advance a wide range of language tasks~\citep{devlin2019bert, radford2019language, zhang2022opt, gemmateam2024gemma, touvron2023llama}, but training on text alone leaves them weak in visual commonsense. Vision–Language Models (VLMs) jointly train on images and text~\citep{alayrac2022flamingo, liu2023llava, liu2023improvedllava, li2023blip2, dai2023instructblip, cha2024honeybee}, improving visually grounded abilities such as VQA~\citep{zhang2022visual, xia2023imagenetvc, li2023vec, jin2024winoviz}. However, they require heavy multimodal training and can reduce non-visual language performance, making rapid adaptation to new LLMs costly. The question is how to add robust visual knowledge to text-only models efficiently.

Visually-augmented LLMs (VaLMs) inject visual signals into pretrained LLMs without full multimodal retraining~\citep{wang2023visuallyaugmented, guo2023visuallyaugmented, zhang2022visual, cui2024moremultimodalretrievalaugmented, tan2020vokenization}, improving visual commonsense and sometimes even non-visual tasks~\citep{Zhang2023TowardsVA, lu2022imaginationaugmented, tang2023learning, zhang2022visual, yang2022zlavi, huang2023languageneedaligningperception}. Yet many methods fuse modalities early and rely on a single image, which can disturb LLM behavior and introduce noise and biases.

\begin{figure}[t!]
  \centering
  \includegraphics[width=1.0\linewidth]{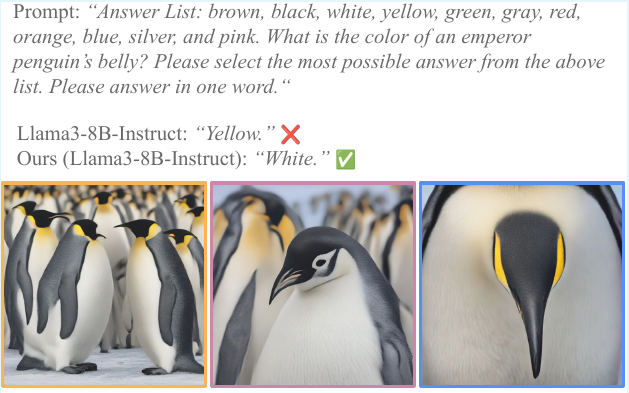}
    \caption{\method at inference. While a text-only LLM fails on penguin belly color, \method generates diverse visual evidence that is late-fused and aggregated to correct the prediction and derive a grounded output.}
  \vspace{-0.3cm}
  \label{fig:teaser}
\end{figure}

We propose a VaLM-style approach that addresses these issues through two key components. First, we introduce a \emph{late fusion} architecture that integrates projected visual features with an LLM only at the final stage of prediction. Given an image and caption during training, a pre-trained multimodal encoder maps the image to a joint image–text space, which is then projected to a sequence of pseudo-text embeddings $z^v_{1:n}$. In parallel, the input text is processed by a pre-trained LLM to produce token embeddings $z^x_{1:k}$. A late-stage attention-like mechanism allows $z^x_{1:k}$ to attend to $z^v_{1:n}$ once, immediately before prediction, rather than feeding visual tokens into the LLM stack. This design keeps the LLM focused on language while enabling access to visual information when helpful.

Second, at inference, we introduce \emph{multi-image} evidence (see Figure~\ref{fig:teaser}). Since paired images are not available at test time, we generate $k$ images from the input text using a \emph{distilled} text-to-image generator with batched, parallel sampling to minimize overhead. Each generated image is processed through the late-fusion module to produce a probability distribution. We also compute a text-only distribution. We then aggregate these $k\!+\!1$ distributions using entropy-aware weighting, allowing confident predictions to dominate while preserving the text-only path when visual input is unhelpful.

We evaluate on object commonsense~\citep{wang2023visuallyaugmented}, visual commonsense (ImageNetVC~\citealp{xia2023imagenetvc}), and standard language benchmarks~\citep{dubey2024llama, touvron2023llama, mosaicml2023introducing, almazrouei2023falcon}. Our method substantially outperforms LLMs and prior VaLMs on visual commonsense, matches VLMs on vision-heavy tasks, and, when applied to strong LLMs such as LLaMA~3, also improves text-only performance, while adding only modest test-time overhead due to lightweight, batched generation.

\section{Related Work}

\textbf{Large Language and Vision Models.} LLMs achieve strong performance on text-based reasoning but struggle with visual understanding~\citep{gemmateam2024gemma}. Vision–Language Models (VLMs)~\cite{liu2023improvedllava, dai2023instructblip, cha2024honeybee} address this gap and excel at multimodal tasks such as VQA, image captioning, and visual commonsense reasoning~\cite{xia2023imagenetvc, li2023vec, jin2024winoviz}, yet often degrade on purely textual commonsense reasoning.

\textbf{Visually-Augmented Language Models.} VaLMs augment text-only LMs with visual inputs. Some retrieve related images and feed them to the LM~\citep{tan2020vokenization, lu2022imaginationaugmented, wang2023visuallyaugmented}, while others distill visual knowledge from multimodal models such as CLIP~\citep{radford2021learning} or BLIP-2~\citep{li2023blip2} into LMs~\citep{tang2021vidlankd, Zhang2023TowardsVA, guo2023visuallyaugmented, li2023vec, cui2024moremultimodalretrievalaugmented}. Diffusion-based approaches like Z-LaVi~\citep{yang2022zlavi} generate visuals for possible label predictions, whereas our method generates images directly from the input text. LiVE~\citep{tang2023learning} adds a vision–text fusion layer into the LM, and iNLG~\citep{zhu2023visualize} uses visual prefixes to guide generation. In contrast, we keep the LLM unchanged, apply late fusion between its output and an image encoding, and aggregate predictions from multiple generated images via simple averaging, enabling diverse visual experts to guide towards confident outputs.

\begin{figure*}[t!]
  \centering
  \includegraphics[width=0.94\linewidth]{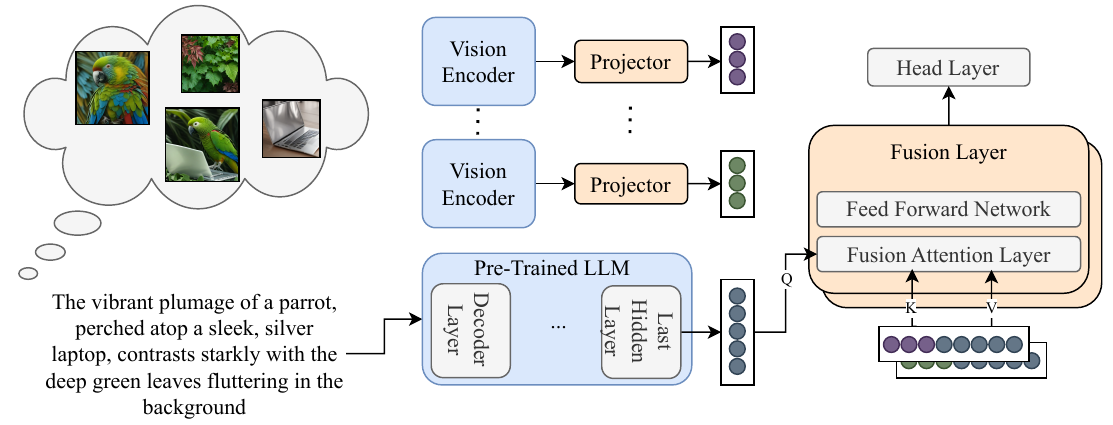}
    \caption{\textbf{Overview of \method.} Multiple images are generated from the input text and independently encoded by a frozen vision encoder, then projected to pseudo-text tokens. In parallel, the text is processed by a frozen pre-trained LLM. A trainable late-fusion attention layer allows the LLM's final text representations to attend to the projected visual tokens, combining both modalities before the prediction head. Blue: frozen; orange: trainable.}
  \vspace{-0.5cm}
  \label{fig:arch}
\end{figure*}

\section{Method}

Our method, \method\ (\textbf{La}te \textbf{M}ulti-\textbf{I}mage fusion), enhances LMs with visual cues to improve object and visual commonsense reasoning while preserving text-only performance. We train with (i) image–caption pairs and (ii) text with a synthetically generated image from a text-to-image model. At inference, we generate $k$ images for a prompt, process each with the model, and aggregate their predictions using confidence-based weighting.

\subsection{Visually Enhanced Language Model}
\label{sec:train}

The architecture, illustrated in Fig.~\ref{fig:arch}, consists of four components: a frozen pre-trained LLM, a frozen pre-trained vision encoder, a trainable Visual Token Projector (VTP), and a trainable Late Fusion Attention Layer (LFAL). 

Given an image $v$ and caption $x = (x_{(1)}, \dots, x_{(n_x)})$, the training objective is:
\begin{align}
    \max_{\theta} \ \log P_\theta\big(x_{(t)} \mid x_{(<t)}, v\big).
\end{align}

The vision encoder extracts patch features $z^v \in \mathbb{R}^{n_v \times d_v}$, which the VTP maps to pseudo-text embeddings via  
\[
    u^v = W_1\,\sigma(W_2 z^v), \quad u^v \in \mathbb{R}^{n_v \times d_x}.
\]
The LFAL fuses $u^v$ with text embeddings $z^x_{(<t)}$ by setting $K=V=[u^v; z^x_{(<t)}]$ and $Q=z^x_{(<t)}$, enabling text tokens to attend once to visual tokens before projecting to vocabulary logits.

\subsection{Visually Driven Inference}
\label{sec:inference}

Since paired images are unavailable at inference, we generate $k$ images $\{v_i\}$ from the prompt using a distilled text-to-image generator with batched, parallel sampling. Each image yields a distribution $p_i = P_\theta(x_t \mid x_{<t}, v_i)$, and we also compute the text-only distribution $p_0 = P_\theta(x_t \mid x_{<t})$.

We ensemble the image-based predictions $p_{\text{img}} = \frac{1}{k} \sum_{i=1}^k p_i,$ then weight them by a normalized CLIP score $f(\bar{x}_i, v_i)$:
\begin{align}
    p_{\text{final}} &= \sum_{i=1}^k f(\bar{x}_i, v_i)\,p_i \notag \\
    &\quad + \big(1 - f(\bar{x}_i, v_i)\big)\,p_0.
\end{align}
High-alignment images are trusted more, while low-alignment ones fall back to the text-only LLM.

\section{Results}

\begin{table}[t!]
  \centering
  \resizebox{\linewidth}{!}{
    \begin{tabular}{lcccc}
      \toprule
      \textbf{Model} & \textbf{Mem.~Color} & \textbf{Color~Terms} & \textbf{Obj.~Shape} & \textbf{Rel.~Size} \\
      \midrule
      \multicolumn{5}{c}{\textbf{Masked Language Models}} \\
      \midrule
      BERT & 31.6  & 30.7 & 28.1 & 38.1 \\
      BERT (FT) & 33.9  & 31.5 & 21.5 & 35.7 \\
      Vokenization$^*$ & 14.2 & 20.0 & 43.2 & 72.4 \\
      X-adapter$^*$ & 64.1 & 60.0 & - & - \\ 
      \midrule
      \method$^\ddagger$ & \textbf{74.5} & \textbf{72.5} & \textbf{67.3} & \textbf{78.4} \\
      \midrule
      \multicolumn{5}{c}{\textbf{Causal Language Models}} \\
      \midrule
      GPT-2 & 32.4  & 34.6 & 44.5 & 43.1 \\
      GPT-2 (FT) & 33.3 & 34.9 & 39.3 & 38.2 \\
      LIVE$^\ddagger$  & 49.6 & 46.7 & 41.5 & 66.7 \\
      % iNLG$^\ddagger$ & 48.6 & 44.8 & 39.5 & 51.1 \\
      Z-LaVI$^*$$^\dagger$ & 50.4 & 49.2 & 64.4 & 76.8 \\
      VaLM$^*$ ($k=4$) & 54.0 & 52.7 & 62.8 & 85.0 \\
      VaLM$^*$ ($k=8$) & 58.6 & 50.2 & 59.4 & 62.4 \\ 
      \midrule
      \method$^\ddagger$ & \textbf{72.5} & \textbf{69.2} & \textbf{66.8} & \textbf{85.5} \\
      \bottomrule
    \end{tabular}}
    \caption{Object commonsense results.
    \( ^*\) retrieves images, \( ^\ddagger\) generates images during inference.}
  \label{tab:vc_benchmark}
\end{table}

We first evaluate \method on object commonsense, comparing against prior visually-augmented LMs and ablating our contributions. We then scale to larger models and evaluate on visual commonsense, reasoning, and reading comprehension. Next, we compare \method against a text-only baseline with matched inference-time budget. Finally, we analyze the effect of the number of generated images, compare integrating generated images against integrating multimodal embeddings, and ablate the CLIP-based fusion strategy. Additional analyses on generation vs.\ retrieval are provided in Sec.~\ref{app:gen_vs_ret}.
\paragraph{Evaluation Benchmarks.} We follow the zero-shot benchmark of~\citet{wang2023visuallyaugmented} for QA on object color, shape, and size, using Memory Color~\citep{norlund2021transferring} and Color Terms~\citep{Bruni2012DistributionalSI} for color, ViComTe~\citep{zhang2022visual} for shape, and the size dataset of~\citet{bagherinezhad2016elephants}, adhering to~\citet{wang2023visuallyaugmented} guidelines.

\vspace{-0.1cm}
\paragraph{Comparison to Visually Augmented.}

Following the baselines, we start by studying weaker LMs, focusing on masked LMs (BERT) and causal LMs (GPT-2), both lacking visual commonsense (e.g., failing to answer \emph{What is the color of a banana?}). For GPT-2 we use zero-shot accuracy; for BERT we follow \citet{Zhang2023TowardsVA}, masking after the last word and predicting the masked token. To ensure a fair comparison, in this experiment \method\ is trained only on the Visual Genome (VG) dataset~\citep{krishna2016visual}, which contains an equal or smaller multimodal data than the baselines. Further implementation and baseline details are provided in the Appendix, Secs.~\ref{app:implementation}, \ref{app:fairness}.

Table~\ref{tab:vc_benchmark} shows that \method\ substantially improves performance across all tasks, with only a marginal gain over VaLM on Relative Size (85.0 vs.\ 85.5). 
We attribute these improvements to integrating multiple generated images via late fusion, which is more robust than baselines that rely on a single image, early fusion, or simple probability summation.

\begin{table}[t!]
  \centering
  \resizebox{\linewidth}{!}{%
    \begin{tabular}{lcccc}
      \toprule
      \textbf{Method} & \textbf{Mem. Color} & \textbf{Color Terms} & \textbf{Obj. Shape} & \textbf{Rel. Size} \\
      \midrule
      GPT-2 (Base) & 32.4  & 34.6 & 44.5 & 43.1 \\
      E-F. & 49.1 & 45.3 & 40.3 & 70.1 \\
      E-F. + M & 55.5 & 52.1 & 41.2 & 75.5 \\
      I-F. & 62.8 & 59.3 & 60.0 & 77.2 \\
      I-F. + M & 69.7 & 67.8 & 63.0 & 81.1 \\
      L-F. & 65.1 & 62.2 & 63.5 & 80.2 \\
      L-F. + M (Ours) & \textbf{72.5} & \textbf{69.2} & \textbf{66.8} & \textbf{85.5} \\
      \bottomrule
    \end{tabular}
  }  
  \caption{Ablation studies on fusion strategies and multi-image generation. 
  Abbreviations: E-F.=Early Fusion, I-F.=Intermediate Fusion, L-F.=Late Fusion, M=Multi-Image Generation.}
  \label{tab:ablation}
\end{table}

To validate our assumptions about the contributing factors of our proposed components, we conduct ablation experiments. Table~\ref{tab:ablation} reports results for seven configurations of the GPT-2 model: early fusion following iNLG~\citep{zhu2023visualize} and LiVE~\citep{tang2023learning}, early fusion with multi-image generation, intermediate fusion as in VaLM~\citep{wang2023visuallyaugmented}, intermediate fusion with multi-image generation, late fusion alone, and our complete approach combining late fusion with multi-image generation.

Results suggest that both architectural choices substantially improve performance, and that removing them reduces our method to levels comparable with the baselines. Multi-image generation consistently yields gains across all fusion strategies, particularly on color and relative size reasoning. Late fusion outperforms early and intermediate fusion, especially on shape-related tasks. The combination of late fusion and multi-image generation achieves the best overall performance.

\vspace{-0.1cm}
\paragraph{Comparison to VLMs.} 
Next, we evaluate \method\ on visual commonsense, commonsense reasoning, and reading comprehension using more advanced models, following the evaluation settings from~\citet{touvron2023llama}. Detailed benchmarks and implementation settings are provided in Appendix, Secs.~\ref{app:benchmarks} and~\ref{app:implementation}. Results provided in Tab.~\ref{tab:text_evals}, \method\ consistently improves LMs of all sizes: small (GPT-2), medium (Gemma-2B), and large (Vicuna-7B-V1.5, Llama3-8B, Llama3-8B-Instruct). 
Unlike VLMs such as InstructBLIP and Llava-Next, which often improve visual commonsense at the cost of text-task performance, \method\ enhances visual commonsense while also maintaining or improving text-based results.  This highlights a key strength of late fusion: \textit{it adds visual capability without sacrificing language reasoning}.

\begin{figure}[t!]
    \centering
    \includegraphics[width=0.99\linewidth]{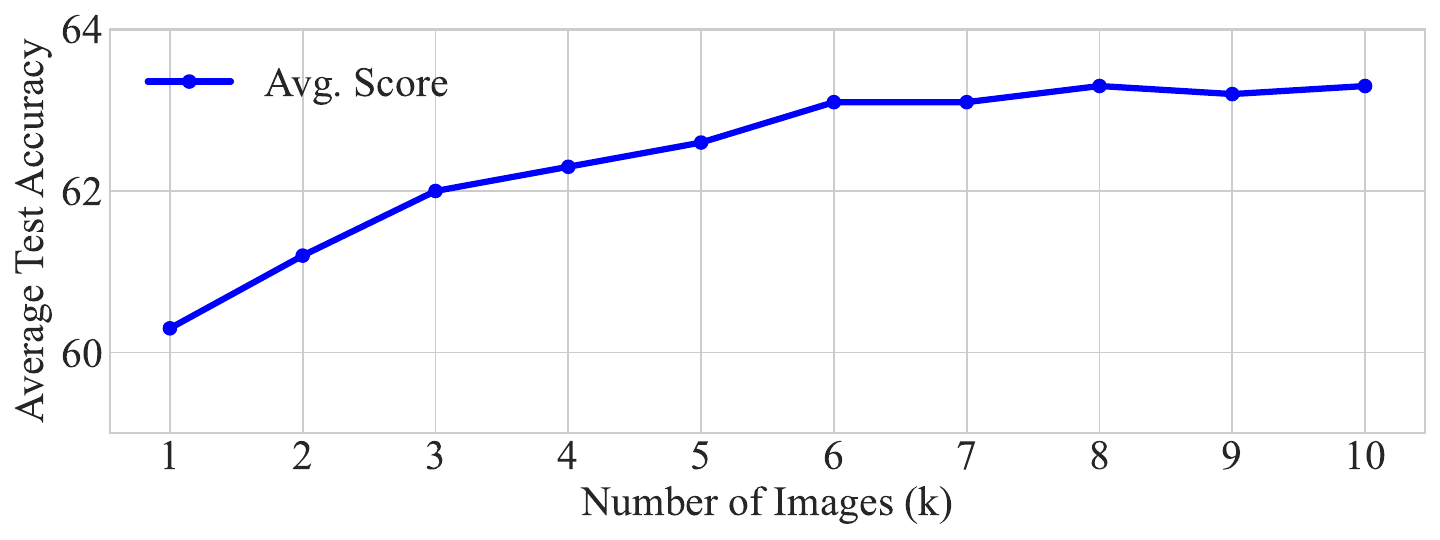}
    \caption{Average performance as a function of the number of generated images.
}
    \label{fig:effective_k}
\end{figure}

\paragraph{Inference-Time Compute Baseline.}
Since \method\ introduces additional inference cost through image generation, a natural question is whether comparable gains could be achieved by simply spending more compute on text-only decoding. To test this, we compare against a Best-of-$N$ strategy: the model generates $N$ independent completions and selects the one with the highest average token log-likelihood, with $N$ calibrated so that total runtime matches \method\ ($k=6$) on Gemma-2B.
As shown in Tab.~\ref{tab:best_of_n}, Best-of-$N$ yields modest gains on commonsense reasoning and reading comprehension but fails to close the visual commonsense gap, confirming that \method's improvements stem from grounded visual evidence rather than additional compute alone.

\begin{table}[t!]
  \centering
  \resizebox{0.75\linewidth}{!}{
    \setlength{\tabcolsep}{6pt}
    \begin{tabular}{lccc}
      \toprule
      \textbf{Method} & \textbf{VC} & \textbf{CR} & \textbf{RC} \\
      \midrule
      Gemma-2B & 45.6 & 63.8 & 48.8 \\
      Gemma-2B (Best-of-$N$) & 47.8 & 64.8 & 49.0 \\
      \method (Gemma-2B) & \textbf{50.1} & \textbf{65.1} & \textbf{48.9} \\
      \bottomrule
    \end{tabular}}
  \caption{Comparison of \method\ against Best-of-$N$ sampling with matched inference runtime on Gemma-2B.}
  \vspace{-0.2cm}
  \label{tab:best_of_n}
\end{table}

\paragraph{Effect of $k$.} In Figure~\ref{fig:effective_k}, we analyze the effect of the number of generated images during inference, varying $k$ from 1 to 10 on a validation set drawn from Color~\citep{xia2023imagenetvc}, PIQA~\citep{bisk2019piqa}, and BoolQ~\citep{clark2019boolq}. Performance improves as $k$ increases, saturating around $k\!\approx\!6$. Since image generation at inference is costly, we note that even $k=3$ yields improvements.

\begin{table}[t!]
  \centering
  \renewcommand{\arraystretch}{1.0}
  \resizebox{\linewidth}{!}{
    \begin{tabular}{llcccc}
      \toprule
      \textbf{Model} & \textbf{Base} & \textbf{VC} & \textbf{CR} & \textbf{RC} & \textbf{Avg.} \\
      \midrule
      \multicolumn{6}{c}{\emph{Small-Scale Models}} \\
      \midrule
      GPT-2 & - & 30.3 & 46.1 & 30.5 & 35.6 \\
      \method & GPT-2 & \textbf{38.6} & \textbf{46.7} & \textbf{32.2} & \textbf{39.2} \\
      \midrule
      \multicolumn{6}{c}{\emph{Mid-Scale Models}} \\
      \midrule
      Gemma-2B & - & 45.6 & 63.8 & 48.8 & 52.7 \\
      \method & Gemma-2B & \textbf{50.1} & \textbf{65.1} & \textbf{48.9} & \textbf{54.7} \\
        \midrule
      \multicolumn{6}{c}{\emph{Large-Scale Models}} \\
      \midrule
      Vicuna-7B & - & 45.1 & 57.6 & 57.5 & 53.4 \\
      InstructBLIP$^*$ & Vicuna-7B & 50.1 & 52.6 & 53.6 & 52.1 \\
      Llava-Next$^*$ & Vicuna-7B & \textbf{50.3} & 54.5 & 54.7 & 53.1 \\
      \method & Vicuna-7B & 48.6 & \textbf{58.8} & \textbf{57.9} & \textbf{55.1} \\
      \midrule
      Llama3-8B & - & 52.0 & 72.0 & 57.9 & 60.6 \\
      \method & Llama3-8B & \textbf{55.0} & \textbf{72.9} & \textbf{58.0} & \textbf{62.0} \\
      \midrule
      \multicolumn{6}{c}{\emph{Large-Scale Instruct Models}} \\
      \midrule
      Llama3-8B-Instruct & - & 53.0 & 71.6 & 59.2 & 61.2 \\
      Llava-Next$^*$ & Llama3-8B-Inst. & \textbf{56.5} & 70.8 & 54.8 & 60.7 \\
      \method & Llama3-8B-Inst. & 55.6 & \textbf{71.7} & \textbf{60.9} & \textbf{62.7} \\
      \midrule
      Qwen-2.5-7B-Instruct & - & 54.8 & 75.6 & 63.0 & 64.4 \\
      Qwen2-VL-7B-Instruct$^*$ & Qwen-2.5-7B-Inst. & \textbf{59.0} & 74.4 & 57.9 & 63.7 \\
      \method & Qwen-2.5-7B-Inst. & 57.8 & \textbf{75.9} & \textbf{64.4} & \textbf{66.0} \\

      \bottomrule
    \end{tabular}
  }  
  \caption{Results on visual commonsense (VC), commonsense reasoning (CR), and reading comprehension (RC). 
  $^*$ = VLM trained on large-scale image–text data.}
  \label{tab:text_evals}
\end{table}

\paragraph{Images vs. Embeddings.}
One might argue that multimodal representations could inject visual information instead of synthetic images~\citep{guo2023visuallyaugmented}. 
This raises the question: \emph{do synthetically generated images hold more information than multimodal text representations?}
To address this, we compare \method (Gemma-2B) against CLIP text embeddings, maintaining identical architecture, datasets, and implementation while replacing only the visual representations. Following~\citet{guo2023visuallyaugmented}, we extract noun entities via part-of-speech tagging before CLIP encoding, as full-text embeddings yielded poor performance. We evaluate under the settings in~\ref{app:benchmarks}, reporting results in Tab.~\ref{tab:clip_text_embedding}. While CLIP embeddings provide a computationally cheaper alternative that still improves over baseline, generated images outperform both CLIP embeddings and baseline across all tasks.

\begin{table}[t!]
  \centering
  \renewcommand{\arraystretch}{0.9}
  \resizebox{0.8\linewidth}{!}{
      \setlength{\tabcolsep}{6pt}
      \begin{tabular}{lcccc}
        \toprule
        Method & VC & CR & RC & \makecell{Time (ms)} \\
        \midrule
        Gemma-2B & 45.6 & 63.8 & 48.8 & \textbf{20} \\
        CLIP Embedding & \underline{47.9} & \underline{64.7} & \underline{48.9} & \underline{58} \\
        Generated Images & \textbf{50.1} & \textbf{65.1} & \textbf{48.9} & 389 \\
        \bottomrule
      \end{tabular}}
  \caption{Performance comparison of image generation, CLIP text embedding, and baseline (Gemma-2B).}
  \label{tab:clip_text_embedding}
  \vspace{-0.4cm}
\end{table}

\paragraph{CLIP-Fusion Ablation.}
Lastly, we ablate the CLIP-based fusion strategy described in Sec.~\ref{sec:inference} by comparing it against purely confidence-based aggregation methods. Specifically, we evaluate: (i)~Single Image, using a single generated image without aggregation; (ii)~Average Logits, averaging the prediction distributions across all $k$ generated images; and (iii)~Max Confidence, selecting the prediction with the lowest entropy among the $k$ image-conditioned distributions. All variants use Gemma-2B with $k=6$ and are evaluated on Color, PIQA, and BoolQ.
Results are reported in Tab.~\ref{tab:clip_fusion}. CLIP-fusion consistently outperforms all confidence-based alternatives.

\begin{table}[t!]
  \centering
  \resizebox{0.8\linewidth}{!}{
    \setlength{\tabcolsep}{6pt}
    \begin{tabular}{lccc}
      \toprule
      \textbf{Method} & \textbf{Color} & \textbf{PIQA} & \textbf{BoolQ} \\
      \midrule
      Single Image & 40.8 & 76.1 & 66.1 \\
      Average Logits & 44.6 & 76.5 & 66.5 \\
      Max Confidence & 43.0 & 76.1 & 66.9 \\
      \method (CLIP-fusion) & \textbf{45.4} & \textbf{77.7} & \textbf{67.0} \\
      \bottomrule
    \end{tabular}}
  \caption{Ablation of aggregation strategies. CLIP-fusion outperforms purely confidence-based alternatives.}
  \vspace{-0.4cm}
  \label{tab:clip_fusion}
\end{table}

\section{Qualitative Analysis and Failure Cases}

We provide representative examples using Llama-3 to illustrate the behavior of \method.

\paragraph{Success case.} \emph{How many humps does a Bactrian camel have?} Llama-3 predicts \emph{one}, confusing Dromedary and Bactrian species. \method\ generates images of two-humped camels, correcting the prediction to \emph{two}.

\paragraph{Robustness to negation.} \emph{Which color is not on a stop sign?} The generator produces a red stop sign, which is misleading under negation. However, the low CLIP alignment score suppresses the visual path, and \method\ falls back to the text-only prediction, correctly outputting \emph{blue}.

\paragraph{Failure case.} \emph{What material holds the Sword of Damocles?} Llama-3 predicts \emph{a thin rope}. \method\ generates images depicting a metal chain, a visually plausible but incorrect depiction (the correct answer is \emph{a single horse hair}). The high alignment score causes the visual path to override the text prior. Such failures are more likely for abstract or legendary concepts where text-to-image generators lack faithful grounding.

\section{Conclusions}
We introduce \method, a framework for enhancing the visual commonsense capabilities of LLMs without compromising their text reasoning. Unlike prior work, we focus on two overlooked but crucial aspects of visual augmentation: (i) leveraging multiple generated images to capture diverse visual evidence, and (ii) applying late fusion over text and visual features for robust integration. Comprehensive evaluation shows that both components are essential, with \method achieving strong gains on visual commonsense tasks while preserving text-based performance.

\section{Limitations}
Visually augmented techniques incur additional cost, as image generation is slower than text decoding. For instance, Gemma-2B requires $\sim$20\,ms per token, while our approach adds $\sim$50\,ms per image. Efficiency can be improved through parallelized image generation, which bounds latency, or by applying visual conditioning as initial context or selectively during decoding.

Nevertheless, this trade-off reflects a broader trend: scaling test-time compute enhances output quality~\citep{wei2023chainofthoughtpromptingelicitsreasoning, yao2023treethoughtsdeliberateproblem, snell2024scalingllmtesttimecompute}, consistent with the ``no-free-lunch'' principle that stronger performance demands longer runtime. In this light, visually augmented reasoning represents a principled form of test-time scaling, likely to become more natural within agentic frameworks. Thus, we view it not as a stopgap but as a practical and forward-looking direction for visual reasoning.

Finally, we have not yet evaluated \method on the latest reasoning LLMs~\citep{deepseekai2025deepseekr1incentivizingreasoningcapability} or scaled training beyond 8B parameters due to computational constraints. We believe that reasoning with augmented images has the potential to advance state-of-the-art performance further.

\section{Ethical Considerations}

Our method inherits risks from its component models: the base LLM, vision encoder (CLIP), and text-to-image generator (SDXL-turbo). Key considerations include:

\textbf{Hallucinations and Factual Grounding.} Both the LLM and text-to-image generator may produce outputs not grounded in facts. Generated images may depict incorrect visual information, potentially compounding reasoning errors.

\textbf{Bias Propagation.} Biases present in the pre-trained LLM, CLIP encoder, and image generator can propagate through our pipeline, leading to unfair or stereotypical visual representations that influence final predictions.

\textbf{Computational Cost.} While our model primarily uses pre-trained foundation models as part of our model design and only adapts a lightweight vision projector and a fusion layer, training such pre-trained models requires significant energy consumption. Further, inference time queries, which are performed many times, may be costly.

\bibliography{custom}

\appendix
\label{sec:appendix}

\section{Experimental Settings}

\subsection{Implementation Details}

\label{app:implementation}

We use CLIP-ViTB/32~\citep{radford2021learning} as the vision encoder and for computing text-image alignment scores, with SDXL-turbo~\citep{sauer2023adversarialdiffusiondistillation} for image generation. Training employed a two-stage pipeline on four A100 GPUs: 40K iterations (batch size 256, lr $5 \times 10^{-4}$) followed by 10K fine-tuning iterations (batch size 128, lr $5 \times 10^{-5}$) using AdamW. Only LFAL and VTP were trained while other components remained frozen. Training required 192 GPU hours for Llama-3, 90 for Gemma-2B/OPT-2.7B, and 50 for GPT-2. 

We trained on Visual Genome Regions (5.4M images)~\citep{krishna2016visual}, Laion-220K~\citep{LAION_LVIS_220}, and 2\% of Wikitext-103~\citep{merity2016pointer}. For Wikipedia texts, we synthetically generated corresponding images to simulate inference conditions. During inference, we generate 6 images per sample unless specified otherwise.

\subsection{VaLM Baseline Details}
\label{app:fairness}
% \paragraph{VaLM Baseline Details.}
We compared our method with several VaLMs designed to improve visual commonsense: Vokenization~\citep{tan2020vokenization} and X-adapter~\citep{Zhang2023TowardsVA} (BERT-based); Z-LaVI~\citep{yang2022zlavi} (GPT-neo-1.3B~\citep{gao2020pile}); LIVE~\citep{tang2023learning} (BART-based~\citep{lewis2019bart}); and VaLM~\citep{wang2023visuallyaugmented} (GPT-2). We also compared against pure LMs (BERT and GPT-2) and their fine-tuned versions trained on the same data without images.

For fair comparison, in this comparison, we trained our method only on the VG dataset. All baselines except VaLM either trained on VG during pretraining or retrieved VG images during inference. Specifically, Vokenization and X-adapter use COCO~\citep{lin2015microsoftcococommonobjects} and VG; LIVE incorporates COCO, VG, CC3M~\citep{sharma-etal-2018-conceptual}, and Flickr30k~\citep{plummer2016flickr30k}. Z-LaVI, a zero-shot model, was employed with VG and Bing Image Search collections. VaLM trains GPT-2 from scratch and lacks publicly available weights, so we report their published results.

For the binary Relative Size test, GPT-2, BERT, Vokenization, and LIVE exhibited strong yes/no bias. We addressed this by fine-tuning open-weight models on 3,200 yes/no questions from ViComTe size~\citep{zhang2022visual} for three epochs with learning rate $5 \times 10^{-5}$.

\subsection{Visual Commonsense, Commonsense Reasoning, and Reading Comprehension Benchmarks}
\label{app:benchmarks}
We evaluate \method\ across three benchmark categories following~\citet{touvron2023llama}. For visual commonsense, we use ImageNetVC~\citep{xia2023imagenetvc}. For Commonsense Reasoning, we evaluate on PIQA~\citep{bisk2019piqa}, SIQA~\citep{sap2019socialiqa}, HellaSwag~\citep{zellers2019hellaswag}, WinoGrande~\citep{sakaguchi2021winogrande}, ARC easy and challenge~\citep{clark2018think}, OpenBookQA~\citep{mihaylov2018can}, and CommonsenseQA~\citep{talmor2018commonsenseqa}. For Reading Comprehension, we use BoolQ~\citep{clark2019boolq}, SQuAD 2.0~\citep{rajpurkar2018know}, and QuAC~\citep{choi2018quac}. For base models, we measure accuracy by selecting the answer with highest likelihood from candidate sets following~\citet{shwartz2020unsupervised}. For instruct models (Vicuna-7B-V1.5, Llama3-8B-Instruct), we use instruction-style prompts with top-1 accuracy following~\citet{xia2023imagenetvc}. SQuAD and QuAC use exact match scoring per~\citet{ouyang2022training}, while BoolQ employs zero-shot binary selection between yes/no tokens.

\section{Generation vs.\ Retrieval}
\label{app:gen_vs_ret}

Prior VaLM baselines such as VaLM~\citep{wang2023visuallyaugmented} and X-adapter~\citep{Zhang2023TowardsVA} rely on image retrieval. While our main results (Tab.~\ref{tab:vc_benchmark}) already show that \method\ outperforms these methods, we further isolate the effect by replacing our generation module with VaLM's retrieval mechanism, keeping all other components identical. On GPT-2, this substitution reduces Memory Color from 72.5 to 65.5 and Color Terms from 69.2 to 62.8, confirming that generation offers superior input specificity and diversity.

\end{document}